\documentclass[conference]{IEEEtran}
\IEEEoverridecommandlockouts
\usepackage{cite}
\usepackage[colorlinks,linktoc=all]{hyperref}
\usepackage{amsmath,amssymb,amsfonts}
\usepackage{algorithmic}
\usepackage{graphicx}
\usepackage{textcomp}
\usepackage{xcolor}
\definecolor{mred}{RGB}{238, 34, 12}
\definecolor{mgreen}{RGB}{1, 127, 0}
\definecolor{mblue}{RGB}{0, 77, 158}
\usepackage{colortbl}
\usepackage{booktabs}
\usepackage{pifont}
\usepackage{wasysym}
\usepackage{arydshln} 
\newcommand{\mredbf}[1]{\textcolor{mred}{\textbf{#1}}}
\newcommand{\mbluebf}[1]{\textcolor{mblue}{\textbf{#1}}}
\def\BibTeX{{\rm B\kern-.05em{\sc i\kern-.025em b}\kern-.08em
    T\kern-.1667em\lower.7ex\hbox{E}\kern-.125emX}}
\begin{document}

\title{Fine-Grained Human Pose Editing Assessment via Layer-Selective MLLMs}


\author{Ningyu Sun\textsuperscript{1}, Zhaolin Cai\textsuperscript{1}, Zitong Xu\textsuperscript{1}, Peihang Chen\textsuperscript{1}, \\ Huiyu Duan\textsuperscript{1}, Yichao Yan\textsuperscript{1}, Xiongkuo Min\textsuperscript{1}\textsuperscript{*}\thanks{* Corresponding Author}\, Xiaokang Yang\textsuperscript{1}\\
\textsuperscript{1}Institute of Image Communication and Network Engineering, Shanghai Jiao Tong University\\
$\{$jiaotong870810, xuzitong, ph4.66, huiyuduan, yanyichao, minxiongkuo, xkyang$\}$@sjtu.edu.cn
}

\maketitle

\begin{abstract}

Text-guided human pose editing has gained significant traction in AIGC applications. However,it remains plagued by structural anomalies and generative artifacts. Existing evaluation metrics often isolate authenticity detection from quality assessment, failing to provide fine-grained insights into pose-specific inconsistencies. To address these limitations, we introduce HPE-Bench, a specialized benchmark comprising 1,700 standardized samples from 17 state-of-the-art editing models, offering both authenticity labels and multi-dimensional quality scores. Furthermore, we propose a unified framework based on layer-selective multimodal large language models (MLLMs). By employing contrastive LoRA tuning and a novel layer sensitivity analysis (LSA) mechanism, we identify the optimal feature layer for pose evaluation. Our framework achieves superior performance in both authenticity detection and multi-dimensional quality regression, effectively bridging the gap between forensic detection and quality assessment.

\end{abstract}


\section{Introduction}

The rapid evolution of text-guided image editing has transformed content creation, empowering users to manipulate visual semantics through natural language prompts \cite{ACE, ip2p, Magicbrush}. Specifically, human pose editing has emerged as a pivotal research area, allowing for the precise alteration of a subject's action or posture \cite{lu2024coarse}. Distinct from general style transfer, pose editing necessitates strict adherence to structural integrity and spatial geometry. While state-of-the-art diffusion models have achieved milestones, they remain prone to generative artifacts, including limb distortions and unnatural texture blending, which undermine the utility and authenticity of the results.

Pose editing introduces both security concerns, as models may generate unrealistic or harmful manipulations \cite{mvss,pscc}, and quality concerns, since users expect visually natural and instruction-consistent results \cite{lmm4edit}. Forensic methods can identify manipulated content but do not capture perceptual quality \cite{fakeshield}, whereas IQA metrics assess visual fidelity but fail to detect subtle fabricated details \cite{qalign,LPIPS}. In practice, authenticity analysis and quality assessment can reinforce each other, since a deeper understanding of forensic features helps expose unrealistic pose constructions that degrade quality, while quality measurements offer fine-grained feedback that facilitates the detection of subtle, imperceptible manipulations. Whereas these two aspects are inherently interconnected and mutually beneficial, an evaluation framework that integrates forensic detection with quality assessment is meaningful.

\begin{figure}
    \centering
    \includegraphics[width=1\linewidth]{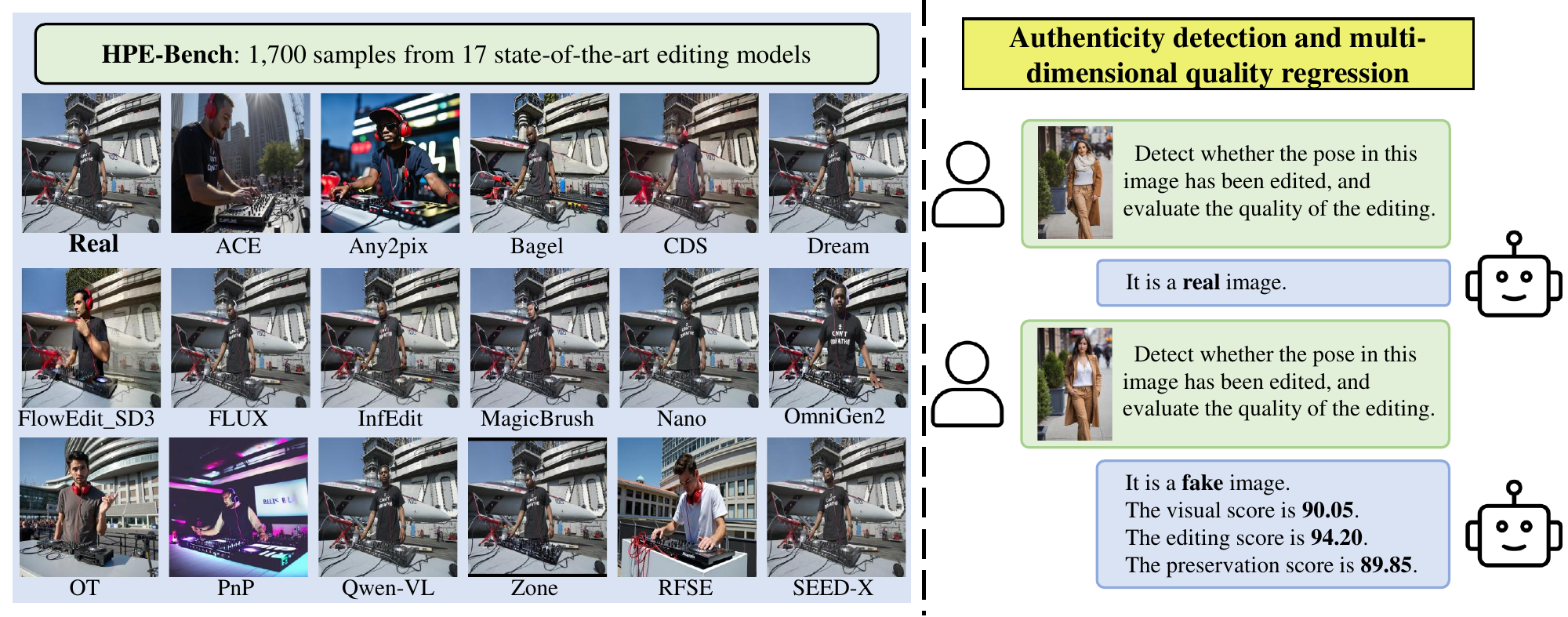}
    \vspace{-5mm}
    \caption{Overview of our constructed HPE-Bench and the proposed evaluation task. HPE-Bench contains 1,700 standardized samples generated by 17 diverse state-of-the-art editing models. Our unified framework performs concurrent authenticity detection and multi-dimensional quality regression, providing scores for visual quality, editing alignment, and attribute preservation.}
    \label{fig:teaser}
    \vspace{-2mm}
\end{figure}

To this end, we introduce \textbf{HPE-Bench} (illustrated in Fig.~\ref{fig:teaser}), a fine-grained benchmark for human pose editing. HPE-Bench comprises 1,700 standardized samples generated by 17 SOTA editing models, spanning both description-based and instruction-based paradigms. Unlike previous datasets that offer only single-dimensional ratings, HPE-Bench provides a rich set of annotations, including authenticity labels for forensic analysis and fine-grained scores in perceptual quality, editing alignment, and attribute preservation. This benchmark serves as a foundational platform for developing evaluation metrics tailored to the complexities of pose synthesis.

Building upon HPE-Bench, we propose a unified framework for authenticity detection and quality assessment, leveraging layer-selective multimodal large language models (MLLMs). Our approach is grounded in the insight that forensic traces are coupled with perceptual quality, the same generative artifacts that expose inauthenticity also degrade visual fidelity. To exploit this correlation, we employ contrastive LoRA tuning to align MLLM representations, maximizing the feature distance between authentic and edited poses. Furthermore, to address the variability of feature distributions, we introduce a layer sensitivity analysis (LSA) mechanism. This module automatically identifies the most informative feature layers by computing statistical metrics, including KL divergence, local discriminant ratio, and feature entropy. Through targeted adaptation, our framework achieves robust performance in both forensic detection and quality assessment tasks.

Our main contributions are as follows:
\begin{itemize}
    \item We construct HPE-Bench, the first fine-grained human pose editing benchmark featuring 1,700 standardized samples from 17 generative models with authenticity labels and multi-dimensional quality annotations.
    \item We propose a unified framework based on layer-selective multimodal large language models, which integrates contrastive LoRA tuning and a novel layer sensitivity analysis mechanism to extract robust discriminative features.
    \item We achieves superior performance in authenticity detection and multi-dimensional quality assessment, effectively bridging the gap between forensic detection and perceptual evaluation in human pose editing.
\end{itemize}

\section{Construction of HPE-Bench}

To support evaluation of fine-grained human pose editing, we construct HPE-Bench, a benchmark emphasizes structural pose transformations, where the human body's geometric configuration is edited while non-target attributes are expected to remain consistent. Each sample is a triplet $(I_{src}, I_{edit}, T)$, where $I_{src}$ denotes the authentic source image, $I_{edit}$ is the pose-edited image generated by editing model, and $T$ represents the pose transformation prompt. This formulation enables both real-fake discriminative learning and instruction-aware quality evaluation within a unified benchmarking framework.

\subsection{Image Collection and Prompt Engineering}

\subsubsection{Source Data Curation}
We collect high-resolution real-world images containing human subjects with clear and well-defined body structures to ensure suitability for pose-driven editing. To ensure precise pose estimation, samples exhibiting severe occlusion, truncation, or ambiguous configurations were excluded. A minimum resolution threshold of $1024 \times 1024$ pixels was enforced to preserve structural fidelity, serving the purpose of high-quality generation and forensic evaluation.

\subsubsection{Prompt Formulation}
For each source image, we generate pose-editing prompts based on images using a MLLM. Each prompt defines the target transformation while enforcing semantic consistency for invariant attributes. To avoid physically implausible motions and ill-defined editing objectives, all prompts are refined through rule-based screening and manual verification. The resulting instruction set encompasses a broad spectrum of modification scenarios, ranging from action transitions to human-object interaction adjustments.

\subsection{Standardized Generation Protocol}

\subsubsection{Generative Models}

We select 17 representative text-guided image editing models spanning both instruction-based and description-based paradigms to cover a broad range of pose-editing behaviors and artifact patterns. Instruction-driven methods include IP2P \cite{ip2p}, MagicBrush \cite{Magicbrush}, Any2Pix \cite{instructany2pix}, ZONE \cite{zone}, HQEdit \cite{HQ}, and ACE++ \cite{ACE}, while description-based approaches include Text2LIVE \cite{Text2live}, EDICT \cite{EDICT}, DDPM \cite{DDPM}, MasaCtrl \cite{Masactrl}, CDS \cite{CDS}, PnP \cite{PnP}, InfEdit \cite{InfEdit}, ReNoise \cite{renoise}, RFSE \cite{RFSE}, FlowEdit (SD3) \cite{Flowedit}, FlowEdit (FLUX) \cite{Flowedit}. These models differ substantially in their generative mechanisms, including diffusion-based generation, rectified flow modeling, and inversion-based control, resulting in diverse pose deformation patterns and artifact distributions.

\subsubsection{Data Synthesis and Analysis}
For each of selected editing models, we apply curated source images and corresponding prompts to generate pose-edited samples, resulting in exactly 100 edited images per model and 1,700 samples in total. To characterize the statistical properties of the generated data, we analyze low-level feature distributions across real and edited images. Results indicate that pose-edited images generally exhibit reduced spatial information but increased colorfulness and contrast, which is consistent with patterns observed in generic manipulation datasets. These deviations further confirm that pose editing introduces detectable structural and textural artifacts, motivating the use of robust representation learning for both detection and quality assessment.

\begin{figure*}[t!]
    \centering
    \includegraphics[width=0.8\linewidth]{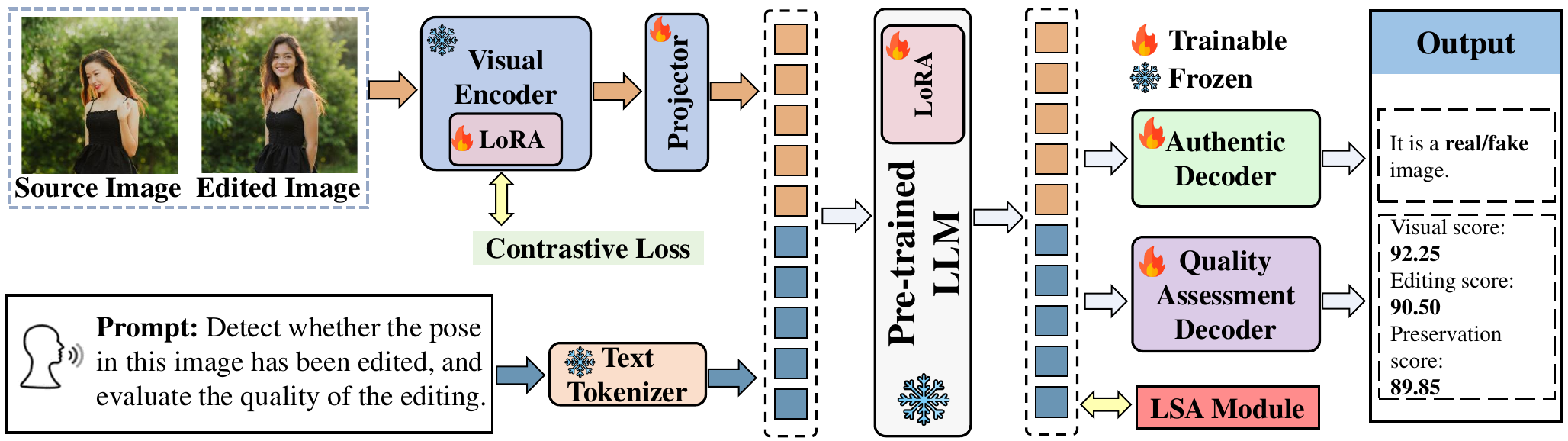}
    \caption{Overview of our proposed framework. The model employs a contrastive LoRA tuning strategy on visual encoder and MLLM to enhance sensitivity to pose-editing artifacts. A layer sensitivity analysis (LSA) module computes statistical metrics to select the optimal intermediate feature layer from the MLLM. Finally, the authentic decoder and quality assessment decoder are utilized for simultaneous authenticity detection and multi-dimensional quality scoring.}
    \vspace{-2mm}
    \label{fig:placeholder}
    \vspace{-2mm}
\end{figure*}

\subsection{Metadata Acquisition and Annotation}

We conduct human evaluation under a controlled setup to obtain reliable subjective annotations for pose editing quality. Annotations are performed using python interface on $3840 \times 2160$ resolution monitor. A total of 20 trained annotators participate in the evaluation process, with samples presented in randomized order and annotation sessions divided into short rounds to mitigate visual fatigue and scoring bias.

Each edited image is annotated following a three-stage protocol consisting of labeling, cross-verification, and expert arbitration to ensure annotation reliability. Annotators provide assessments across three dimensions. Perceptual quality ($S_q$) measures visual fidelity and penalizes visible generative artifacts. Editing alignment ($S_e$) evaluates the semantic consistency between the edited pose and the textual instruction. Attribute preservation ($S_p$) assesses whether non-target regions including background context and identity-related attributes remain unaffected. Beyond these fine-grained metrics (on a 1-5 scale), each sample is also assigned a binary authenticity label to support supervised forensic detection.

\section{Proposed Method}

\subsection{Overview}
We propose a unified framework designed to simultaneously assess the authenticity and quality of pose-edited images. Given a triplet $\{I_{src}, I_{edit}, T\}$, where $I_{src}$ is the source image, $I_{edit}$ is the edited image, and $T$ is the textual instruction, our objective is to predict a binary authenticity label $y_{auth}$ and a multi-dimensional quality vector $\mathbf{y}_{qual} = [s_q, s_e, s_p]^\top$. 
The architecture is built upon a multimodal large language model (MLLM). To overcome the insensitivity of pre-trained MLLM to high-frequency manipulation traces, we introduce contrastive Low-Rank Adaptation (LoRA). Furthermore, to balance forensic discriminability with semantic richness, we propose a layer sensitivity analysis (LSA) mechanism that selects the optimal feature layer. Finally, the selected features feed into specific decoders for detection and regression.

\subsection{Contrastive Visual Tuning}
Standard MLLM prioritize semantic alignment over high-frequency artifact detection. To address this, we apply LoRA tuning, injecting trainable rank decomposition matrices to adapt features for forensic sensitivity without catastrophic forgetting.
To explicitly shape the latent space, we minimize a supervised contrastive loss $\mathcal{L}_{con}$ using triplets of real anchor $f_{src}$, positive real $f_{pos}$, and negative edited $f_{edit}$:
\begin{equation}
\small
    \mathcal{L}_{con} = -\log \frac{\exp(\text{sim}(f_{src}, f_{pos})/\tau)}{\sum_{f \in \{f_{pos}, f_{edit}\}} \exp(\text{sim}(f_{src}, f)/\tau)}
\label{eq:con}
\end{equation}
This objective forces the MLLM to cluster real samples while pushing manipulated ones apart, amplifying feature disparity caused by pose artifacts.

\begin{table*}[h]
\belowrulesep=0pt
\aboverulesep=0pt
\centering
\fontsize{5}{5.5}\selectfont
\setlength{\tabcolsep}{3pt} 
\setlength{\lightrulewidth}{0.5pt}  
\setlength{\heavyrulewidth}{0.6pt}  
\small{
\caption{Comparison results on different manipulation methods. $\spadesuit$ standard CNN/Transformer baselines, $\heartsuit$ deepfake detection methods, $\clubsuit$ AI-generated image detection methods, $\diamondsuit$ multimodal large language models. The fine-tuned results are marked with \raisebox{0.25ex}{\tiny \ding{91}}. The best results are highlighted in \mredbf{red}, and the second-best results are highlighted in \mbluebf{blue}.}
}
\vspace{-1mm}
\resizebox{0.7\textwidth}{!}{\begin{tabular}{l||cccccccccccccccccccc||cc}
\toprule
\noalign{\vspace{1pt}}
Editing Model& \multicolumn{2}{c}{Text2LIVE} & \multicolumn{2}{c}{EDICT} & \multicolumn{2}{c}{IP2P} & \multicolumn{2}{c}{DDPM} & \multicolumn{2}{c}{MasaCtrl} & \multicolumn{2}{c}{CDS} &\multicolumn{2}{c}{MagicBrush} & \multicolumn{2}{c}{PnP}  & \multicolumn{2}{c}{Any2Pix}\\
\noalign{\vspace{-1pt}}
\cmidrule(lr){2-3}
\cmidrule(lr){4-5}
\cmidrule(lr){6-7}
\cmidrule(lr){8-9}
\cmidrule(lr){10-11}
\cmidrule(lr){12-13}
\cmidrule(lr){14-15}
\cmidrule(lr){16-17}
\cmidrule(lr){18-19}
\noalign{\vspace{0.5pt}}
 Model/Metric&Acc$\uparrow$& F1$\uparrow$ &Acc$\uparrow$& F1$\uparrow$ &Acc$\uparrow$& F1$\uparrow$ &Acc$\uparrow$& F1$\uparrow$ &Acc$\uparrow$& F1$\uparrow$ &Acc$\uparrow$& F1$\uparrow$ &Acc$\uparrow$& F1$\uparrow$ &Acc$\uparrow$& F1$\uparrow$ &Acc$\uparrow$& F1$\uparrow$ \\
\midrule
Random Choice& 50.00 & 50.00 & 50.00 &50.00 & 50.00 & 50.00 & 50.00 & 50.00 & 50.00 & 50.00 & 50.00 & 50.00 & 50.00 & 50.00 & 50.00 & 50.00 & 50.00 & 50.00 \\
\hline
$\heartsuit$MVSSNet\raisebox{0.25ex}{\tiny \ding{91}} \cite{mvss} & 57.50 & 66.67 & 62.50 & 75.00 & 65.00 & 78.79 & 67.50 & 82.35 & 60.00 & 70.97 & 56.25 & 64.41 & 61.25 & 73.02 & 65.00 & 78.79 & 60.00 & 70.97 \\
$\heartsuit$PSCCNet\raisebox{0.25ex}{\tiny \ding{91}} \cite{pscc} & 50.00 & 82.35 & 40.00 & 66.67 & 46.25 & 76.92 & 38.75 & 64.41 & 45.00 & 75.00 & 46.25 & 76.92 & 45.00 & 75.00 & 38.75 & 64.41 & 43.75 & 73.02 \\
$\heartsuit$HifiNet\raisebox{0.25ex}{\tiny \ding{91}} \cite{hifi} & 66.25 & 82.35 & 60.00 & 73.02 & 68.75 & 85.71 & 57.50 & 68.85 & 61.25 & 75.00 & 56.25 & 66.67 & 55.00 & 64.41 & 55.00 & 64.41 & 68.75 & 85.71 \\
$\heartsuit$FakeShield\raisebox{0.25ex}{\tiny \ding{91}} \cite{fakeshield}& 72.50 & 84.06 & 77.50 & 90.41 & 73.75 & 85.71 & 77.50 & 90.41 & 72.50 & 84.06 & 71.25 & 82.35 & 71.25 & 82.35 & 73.75 & 85.71 & 75.00 & 87.32 \\
\hline
$\clubsuit$CNNSpot\raisebox{0.25ex}{\tiny \ding{91}} \cite{cnnspot}& 58.75 & 73.02 & 63.75 & 80.60 & 61.25 & 76.92 & 58.75 & 73.02 & 65.00 & 82.35 & 63.75 & 80.60 & 62.50 & 78.79 & 58.75 & 73.02 & 60.00 & 75.00 \\
$\clubsuit$Lagrad\raisebox{0.25ex}{\tiny \ding{91}} \cite{lagrad}& 65.00 & 78.79 & 70.00 & \mbluebf{85.71} & 62.50 & 75.00 & 61.25 & 73.02 & 66.25 & 80.60 & 66.25 & 80.60 & 63.75 & 76.92 & 73.75 & 90.41 & 66.25 & 80.60 \\
$\clubsuit$Univ\raisebox{0.25ex}{\tiny \ding{91}} \cite{univ}& 73.75 & 87.32 & 71.25 & 84.06 & 76.25 & 90.41 & 68.75 & 80.60 & 72.50 & 85.71 & 75.00 & 88.89 & 75.00 & 88.89 & 66.25 & 76.92 & 73.75 & 87.32 \\
$\clubsuit$AIDE\raisebox{0.25ex}{\tiny \ding{91}} \cite{AIDE}& \mbluebf{83.75} & \mbluebf{91.89} & \mbluebf{76.25} & 82.35 & \mbluebf{86.25} & \mbluebf{94.74} & \mbluebf{82.50} & \mbluebf{90.41} & \mbluebf{80.00} & \mbluebf{87.32} & \mbluebf{82.50} & \mbluebf{90.41} & \mbluebf{88.75} & \mbluebf{97.44} & \mbluebf{83.75} & \mbluebf{91.89} & \mbluebf{86.25} & \mbluebf{94.74} \\
\hline
\ding{73}LLaVA-1.6 (7B) \cite{llava}& 53.75 & 66.67 & 56.25 & 70.97 & 51.25 & 62.07 & 51.25 & 62.07 & 62.50 & 80.60 & 57.50 & 73.02 & 51.25 & 62.07 & 53.75 & 66.67 & 62.50 & 80.60 \\
\ding{73}LLaVA-NeXT (8B) \cite{llava-n} & 70.97 & 57.50 & 80.60 & 63.75 & 78.79 & 62.50 & 62.07 & 52.50 & 90.41 & 71.25 & 84.06 & 66.25 & 76.92 & 61.25 & 84.06 & 66.25 & 78.79 & 62.50 \\
\ding{73}mPLUG-Owl3 (7B) \cite{mplug}& 47.50 & 66.67 & 52.50 & 75.00 & 58.75 & 84.06 & 47.50 & 66.67 & 62.50 & 88.89 & 42.50 & 57.14 & 57.50 & 82.35 & 47.50 & 66.67 & 60.00 & 85.71 \\
\ding{73}LLama3.2-Vision (11B) \cite{llama}  & 67.50 & 84.06 & 57.50 & 68.85 & 63.75 & 78.79 & 62.50 & 76.92 & 65.00 & 80.60 & 61.25 & 75.00 & 65.00 & 80.60 & 57.50 & 68.85 & 61.25 & 75.00 \\
\ding{73}MiniCPM-V2.6 (8B) \cite{minicpm}& 55.00 & 66.67 & 63.75 & 80.60 & 62.50 & 78.79 & 52.50 & 62.07 & 66.25 & 84.06 & 50.00 & 57.14 & 58.75 & 73.02 & 58.75 & 73.02 & 55.00 & 66.67 \\
\ding{73}Ovis2.5 (9B) \cite{ovis25}& 42.50 & 54.55 & 43.75 & 57.14 & 47.50 & 64.41 & 48.75 & 66.67 & 47.50 & 64.41 & 53.75 & 75.00 & 53.75 & 75.00 & 40.00 & 49.06 & 53.75 & 75.00 \\
\ding{73}DeepSeekVL2 (small) \cite{deepseekv2}& 58.75 & 75.00 & 57.50 & 73.02 & 56.25 & 70.97 & 61.25 & 78.79 & 61.25 & 78.79 & 61.25 & 78.79 & 53.75 & 66.67 & 62.50 & 80.60 & 61.25 & 78.79 \\
\ding{73}InternVL3.5 (8B) \cite{internvl3_5}& 46.25 & 54.55 & 56.25 & 73.02 & 53.75 & 68.85 & 51.25 & 64.41 & 56.25 & 73.02 & 51.25 & 64.41 & 57.50 & 75.00 & 51.25 & 64.41 & 57.50 & 75.00 \\
\ding{73}Qwen3-VL (8B) \cite{qwen3}& 67.50 & 80.60 & 56.25 & 62.07 & 65.00 & 76.92 & 61.25 & 70.97 & 62.50 & 73.02 & 70.00 & 84.06 & 68.75 & 82.35 & 60.00 & 68.85 & 62.50 & 73.02 \\
\hline
\rowcolor{gray!20}  
Ours & \mredbf{94.74} & \mredbf{90.00} & \mredbf{93.33} & \mredbf{87.50} & \mredbf{96.10} & \mredbf{92.50} & \mredbf{91.89} & \mredbf{85.00} & \mredbf{96.10} & \mredbf{92.50} & \mredbf{98.73} & \mredbf{97.50} & \mredbf{94.74} & \mredbf{90.00} & \mredbf{97.44} & \mredbf{95.00} & \mredbf{96.10} & \mredbf{92.50} \\
\bottomrule
\end{tabular}}

\centering
\resizebox{0.7\textwidth}{!}{\begin{tabular}{l||cccccccccccccccc|cc}
\noalign{\vspace{1.5pt}}
\toprule
\noalign{\vspace{1pt}}
Editing Model& \multicolumn{2}{c}{InfEdit} & \multicolumn{2}{c}{ZONE} & \multicolumn{2}{c}{ReNoise} & \multicolumn{2}{c}{HQEdit} & \multicolumn{2}{c}{RFSE} & \multicolumn{2}{c}{FlowEdit(SD3)} &\multicolumn{2}{c}{FlowEdit(FLUX)} & \multicolumn{2}{c}{ACE++}  & \multicolumn{2}{c}{Overall}\\
\noalign{\vspace{-1pt}}
\cmidrule(lr){2-3}
\cmidrule(lr){4-5}
\cmidrule(lr){6-7}
\cmidrule(lr){8-9}
\cmidrule(lr){10-11}
\cmidrule(lr){12-13}
\cmidrule(lr){14-15}
\cmidrule(lr){16-17}
\cmidrule(lr){18-19}
\noalign{\vspace{0.5pt}}
 Model/Metric&Acc$\uparrow$& F1$\uparrow$ &Acc$\uparrow$& F1$\uparrow$ &Acc$\uparrow$& F1$\uparrow$ &Acc$\uparrow$& F1$\uparrow$ &Acc$\uparrow$& F1$\uparrow$ &Acc$\uparrow$& F1$\uparrow$ &Acc$\uparrow$& F1$\uparrow$ &Acc$\uparrow$& F1$\uparrow$ &Acc$\uparrow$& F1$\uparrow$ \\
\midrule
Random Choice& 50.00 & 50.00 & 50.00 &50.00 & 50.00 & 50.00 & 50.00 & 50.00 & 50.00 & 50.00 & 50.00 & 50.00 & 50.00 & 50.00 & 50.00 & 50.00 & 50.00 & 50.00 \\
\hline
$\heartsuit$MVSSNet\raisebox{0.25ex}{\tiny \ding{91}} \cite{mvss} & 57.50 & 66.67 & 53.75 & 59.65 & 63.75 & 76.92 & 65.00 & 78.79 & 67.50 & 82.35 & 68.75 & 84.06 & 65.00 & 78.79 & 67.50 & 82.35 & 62.57 & 74.74 \\
$\heartsuit$PSCCNet\raisebox{0.25ex}{\tiny \ding{91}} \cite{pscc} & 51.25 & 84.06 & 41.25 & 68.85 & 40.00 & 66.67 & 57.50 & 91.89 & 55.00 & 88.89 & 50.00 & 82.35 & 52.50 & 85.71 & 46.25 & 76.92 & 46.32 & 76.47 \\
$\heartsuit$HifiNet\raisebox{0.25ex}{\tiny \ding{91}} \cite{hifi} & 60.00 & 73.02 & 65.00 & 80.60 & 62.50 & 76.92 & 63.75 & 78.79 & 71.25 & 88.89 & 62.50 & 76.92 & 66.25 & 82.35 & 68.75 & 85.71 & 62.87 & 77.02 \\
$\heartsuit$FakeShield\raisebox{0.25ex}{\tiny \ding{91}} \cite{fakeshield}& 71.25 & 82.35 & 66.25 & 75.00 & 72.50 & 84.06 & 67.50 & 76.92 & 78.75 & 91.89 & 82.50 & 96.10 & 75.00 & 87.32 & 76.25 & 88.89 & 73.82 & 85.58 \\
\hline
$\clubsuit$CNNSpot\raisebox{0.25ex}{\tiny \ding{91}} \cite{cnnspot}& 60.00 & 75.00 & 61.25 & 76.92 & 61.25 & 76.92 & 67.50 & 85.71 & 61.25 & 76.92 & 70.00 & 88.89 & 67.50 & 85.71 & 66.25 & 84.06 & 62.79 & 79.03 \\
$\clubsuit$Lagrad\raisebox{0.25ex}{\tiny \ding{91}} \cite{lagrad} & 68.75 & 84.06 & 57.50 & 66.67 & 66.25 & 80.60 & 67.50 & 82.35 & 68.75 & 84.06 & 65.00 & 78.79 & 72.50 & \mbluebf{88.89} & 68.75 & 84.06 & 66.47 & 80.65 \\
$\clubsuit$Univ\raisebox{0.25ex}{\tiny \ding{91}} \cite{univ}& 63.75 & 73.02 & 68.75 & 80.60 & \mbluebf{77.50} & \mbluebf{91.89} & 72.50 & 85.71 & 76.25 & 90.41 & 71.25 & 84.06 & 73.75 & 87.32 & 76.25 & 90.41 & 72.50 & 85.50 \\
$\clubsuit$AIDE\raisebox{0.25ex}{\tiny \ding{91}} \cite{AIDE}& \mbluebf{86.25} & \mbluebf{94.74} & \mbluebf{78.75} & \mbluebf{85.71} & 76.25 & 82.35 & \mbluebf{83.75} & \mbluebf{91.89} & \mbluebf{86.25} & \mbluebf{94.74} & \mbluebf{85.00} & \mbluebf{93.33} & \mbluebf{77.50} & 84.06 & \mbluebf{83.75} & \mbluebf{91.89} & \mbluebf{82.79} & \mbluebf{90.58} \\
\hline
\ding{73}LLaVA-1.6 (7B) \cite{llava} & 57.50 & 73.02 & 58.75 & 75.00 & 57.50 & 73.02 & 57.50 & 73.02 & 67.50 & 87.32 & 66.25 & 85.71 & 66.25 & 85.71 & 65.00 & 84.06 & 58.60 & 74.21 \\
\ding{73}LLaVA-NeXT (8B) \cite{llava-n} & 85.71 & 67.50 & 68.85 & 56.25 & 87.32 & 68.75 & 90.41 & 71.25 & 88.89 & 70.00 & 85.71 & 67.50 & 88.89 & 70.00 & 93.33 & 73.75 & 82.10 & 65.22 \\
\ding{73}mPLUG-Owl3 (7B) \cite{mplug} & 55.00 & 78.79 & 51.25 & 73.02 & 53.75 & 76.92 & 62.50 & 88.89 & 61.25 & 87.32 & 56.25 & 80.60 & 62.50 & 88.89 & 58.75 & 84.06 & 55.15 & 78.33 \\
\ding{73}LLama3.2-Vision (11B) \cite{llama}  & 63.75 & 78.79 & 63.75 & 78.79 & 66.25 & 82.35 & 63.75 & 78.79 & 70.00 & 87.32 & 73.75 & 91.89 & 67.50 & 84.06 & 75.00 & 93.33 & 65.00 & 80.23 \\
\ding{73}MiniCPM-V2.6 (8B) \cite{minicpm} & 66.25 & 84.06 & 57.50 & 70.97 & 57.50 & 70.97 & 61.25 & 76.92 & 65.00 & 82.35 & 68.75 & 87.32 & 66.25 & 84.06 & 66.25 & 84.06 & 60.66 & 75.46 \\
\ding{73}Ovis2.5 (9B) \cite{ovis25} & 53.75 & 75.00 & 41.25 & 51.85 & 50.00 & 68.85 & 58.75 & 82.35 & 55.00 & 76.92 & 56.25 & 78.79 & 56.25 & 78.79 & 58.75 & 82.35 & 50.66 & 69.18 \\
\ding{73}DeepSeekVL2 (small) \cite{deepseekv2} & 62.50 & 80.60 & 67.50 & 87.32 & 55.00 & 68.85 & 61.25 & 78.79 & 67.50 & 87.32 & 71.25 & 91.89 & 63.75 & 82.35 & 61.25 & 78.79 & 61.40 & 78.67 \\
\ding{73}InternVL3.5 (8B) \cite{internvl3_5} & 53.75 & 68.85 & 61.25 & 80.60 & 62.50 & 82.35 & 55.00 & 70.97 & 53.75 & 68.85 & 55.00 & 70.97 & 60.00 & 78.79 & 57.50 & 75.00 & 55.29 & 71.12 \\
\ding{73}Qwen3-VL (8B) \cite{qwen3} & 66.25 & 78.79 & 65.00 & 76.92 & 70.00 & 84.06 & 61.25 & 70.97 & 73.75 & 88.89 & 65.00 & 76.92 & 67.50 & 80.60 & 72.50 & 87.32 & 65.59 & 77.43 \\
\hline
\rowcolor{gray!20}  
Ours & \mredbf{97.44} & \mredbf{95.00} & \mredbf{90.41} & \mredbf{82.50} & \mredbf{97.44} & \mredbf{95.00} & \mredbf{97.44} & \mredbf{95.00} & \mredbf{97.44} & \mredbf{95.00} & \mredbf{93.33} & \mredbf{87.50} & \mredbf{97.44} & \mredbf{95.00} & \mredbf{93.33} & \mredbf{90.00} & \mredbf{95.50} & \mredbf{91.62} \\
\bottomrule

\end{tabular}}
\label{acc1}
\vspace{-4mm}
\end{table*}

\begin{table}[h]
\small{
\caption{Performance comparisons of quality evaluation methods from perspectives of perceptual \underline{quality}, editing \underline{alignment}, and attribute \underline{preservation}. 
SRCC ($\rho_s$), KRCC ($\rho_k$), and PLCC ($\rho_p$) are reported. 
$\spadesuit$ Traditional FR IQA metrics, $\heartsuit$ traditional NR IQA metrics, $\clubsuit$ deep learning-based FR IQA methods, $\diamondsuit$ deep learning-based NR IQA methods, \ding{72} vision-language methods, \ding{73} LMM-based models. The fine-tuned results are marked with \raisebox{0.5ex}{\scriptsize \ding{91}}. The best results are highlighted in \mredbf{red}, and the second-best results are highlighted in \mbluebf{blue}.}
}
\vspace{-2mm}
\begin{center}
\centering
\centering
\setlength{\tabcolsep}{4pt} 
\resizebox{0.44\textwidth}{!}{
\begin{tabular}{l||ccc|ccc|ccc}
\toprule
Dimensions& \multicolumn{3}{c}{Quality} & \multicolumn{3}{c}{Alignment} & \multicolumn{3}{c}{Preservation} \\
\noalign{\vspace{-2pt}}
\cmidrule(lr){2-4}
\cmidrule(lr){5-7}
\cmidrule(lr){8-10}
\noalign{\vspace{-3pt}}
Methods/Metrics & $\rho_s$$\uparrow$ & $\rho_k$$\uparrow$ & $\rho_p$$\uparrow$ &  $\rho_s$$\uparrow$ & $\rho_k$$\uparrow$ & $\rho_p$$\uparrow$ & $\rho_s$$\uparrow$ & $\rho_k$$\uparrow$ & $\rho_p$$\uparrow$  \\
\hline
$\spadesuit$MSE & 0.0268  & 0.0191  & 0.2215  & 0.2264  & 0.1528  & 0.0060  & 0.4996  & 0.3414  & 0.5415  \\
$\spadesuit$PSNR & 0.0245  & 0.0169  & 0.2228  & 0.2156  & 0.1555  & 0.2609  & 0.4342  & 0.3566  & 0.4508  \\
$\spadesuit$SSIM\cite{SSIM} & 0.0038  & 0.0007  & 0.2207  & 0.1655  & 0.1093  & 0.2206  & 0.4961  & 0.3495  & 0.4519  \\
$\spadesuit$FSIM\cite{FSIM} & 0.0508  & 0.0326  & 0.2242  & 0.2347  & 0.1364  & 0.2607  & 0.5664  & 0.4067  & 0.5636  \\
\hline

$\heartsuit$BIQI\cite{BIQI}  & 0.3182  & 0.1867  & 0.3335  & 0.1180  & 0.0710  & 0.1656  & 0.1623  & 0.0968  & 0.2520  \\
$\heartsuit$DIIVINE\cite{DIIVINE}  & 0.1541  & 0.0918  & 0.3639  & 0.0515  & 0.0310  & 0.1303  & 0.0128  & 0.0081  & 0.1904  \\
$\heartsuit$BRISQUE\cite{BRISQUE}  & 0.3699  & 0.2432  & 0.3868  & 0.1386  & 0.0941  & 0.1497  & 0.1424  & 0.0805  & 0.1945  \\
$\clubsuit$LPIPS \cite{LPIPS} & 0.1903  & 0.1351  & 0.3024  & 0.2066  & 0.1408  & 0.2770  & 0.6867  & 0.5678  & 0.7111  \\
$\clubsuit$ST-LPIPS \cite{STLPIPS} & 0.0054  & 0.0045  & 0.0482  & 0.1961  & 0.1293  & 0.1382  & 0.4575  & 0.3128  & 0.4273  \\
$\clubsuit$CVRKD\raisebox{0.5ex}{\scriptsize \ding{91}} \cite{CVRKD}  & 0.7740  & 0.6128  & \mbluebf{0.8807}  & 0.4403  & 0.3072  & 0.5194  & 0.7827  & 0.6094  & 0.8314  \\
$\clubsuit$AHIQ\raisebox{0.5ex}{\scriptsize \ding{91}}  \cite{AHIQ} & 0.7528  & 0.6367  & 0.7575  & 0.5328  & 0.3928  & 0.5827  & 0.8824  & 0.6578  & 0.8805  \\
\hline

$\diamondsuit$DBCNN\raisebox{0.5ex}{\scriptsize \ding{91}}  \cite{DBCNN} & 0.7294  & 0.6080  & 0.7997  & 0.2959  & 0.2221  & 0.3584  & 0.6074  & 0.4255  & 0.6482  \\
$\diamondsuit$HyperIQA\raisebox{0.5ex}{\scriptsize \ding{91}}  \cite{Hyper}  & 0.7110  & 0.4972  & 0.6919  & 0.2493  & 0.1430  & 0.2348  & 0.2978  & 0.1902  & 0.3063  \\
$\diamondsuit$MANIQA\raisebox{0.5ex}{\scriptsize \ding{91}}  \cite{MANIQA} & 0.7955  & \mbluebf{0.6681}  & 0.8454  & 0.3132  & 0.2584  & 0.3979  & 0.6372  & 0.4653  & 0.6415  \\
$\diamondsuit$TOPIQ\raisebox{0.5ex}{\scriptsize \ding{91}}  \cite{TOPIQ} & 0.8026  & 0.6180  & 0.7919  & 0.3354  & 0.3170  & 0.3552  & 0.5185  & 0.3790  & 0.5638  \\
$\diamondsuit$Q-Align\raisebox{0.5ex}{\scriptsize \ding{91}}  \cite{qalign} & \mbluebf{0.8527}  & 0.5825  & 0.8793  & 0.4518  & 0.3717  & 0.5415  & 0.7119  & 0.4738  & 0.7852  \\
\hline

\ding{72}CLIPScore \cite{clipscore} & 0.2024  & 0.1326  & 0.2417  & 0.2099  & 0.1328  & 0.2355  & 0.2446  & 0.1526  & 0.2597  \\
\ding{72}ImageReward \cite{imagereward} & 0.4288  & 0.2588  & 0.3997  & 0.3079  & 0.2101  & 0.2904  & 0.4304  & 0.2950  & 0.4392  \\
\ding{72}PickScore \cite{pickscore}& 0.2586  & 0.1562  & 0.2683  & 0.3602  & 0.2346  & 0.3416  & 0.1359  & 0.0929  & 0.1977  \\
\ding{72}LLaVAScore \cite{llavascore}& 0.3154  & 0.1934  & 0.3668  & 0.2659  & 0.1832  & 0.2716  & 0.3085  & 0.2015  & 0.4258  \\
\ding{72}VQAScore \cite{vqa}& 0.2796  & 0.2150  & 0.3250  & 0.2587  & 0.1819  & 0.2822  & 0.2267  & 0.1429  & 0.2407  \\
\hline

\ding{73}LLaVA-1.5 (7B) \cite{llava} & 0.1402  & 0.1149  & 0.1189  & 0.2317  & 0.1899  & 0.2391  & 0.0746  & 0.0611  & 0.0642  \\
\ding{73}LLaVA-NeXT (8B) \cite{llava-n} & 0.0818  & 0.0675  & 0.0131  & 0.0089  & 0.0071  & 0.0732  & 0.1591  & 0.1298  & 0.0964  \\
\ding{73}mPLUG-Owl3 (7B) \cite{mplug} & 0.2002  & 0.1383  & 0.0144  & 0.3437  & 0.2406  & 0.0357  & 0.1802  & 0.1289  & 0.1303  \\
\ding{73}LLama3.2-Vision (11B) \cite{llama} & 0.0818  & 0.0675  & 0.0131  & 0.1506  & 0.1213  & 0.0288  & 0.0967  & 0.0785  & 0.1088  \\
\ding{73}MiniCPM-V2.6 (8B) \cite{minicpm} & 0.2706  & 0.2198  & 0.3094  & 0.0710  & 0.0565  & 0.2816  & 0.3581  & 0.2851  & 0.3403  \\
\ding{73}InternVL3 (8B) \cite{internvl3} & 0.4629  & 0.3655  & 0.2437  & 0.1739  & 0.1618  & 0.2445  & 0.3053  & 0.2529  & 0.0002  \\
\ding{73}DeepSeekVL2 (small) \cite{deepseekvl2} & 0.2577  & 0.2104  & 0.2120  & 0.0937  & 0.0769  & 0.0755  & 0.2788  & 0.2266  & 0.1190  \\
\ding{73}Qwen3-VL (7B) \cite{qwen3} & 0.5822  & 0.4427  & 0.5090  & 0.4850  & 0.3549  & 0.4883  & 0.7436  & 0.5638  & 0.7456  \\
\ding{73}LLaVA-NeXT (8B)\raisebox{0.5ex}{\scriptsize \ding{91}} \cite{llava-n} & 0.8161  & 0.6634  & 0.8249  & 0.8437  & 0.6546  & 0.8436  & 0.7821  & 0.6603  & 0.8071  \\
\ding{73}DeepSeekVL2 (small)\raisebox{0.5ex}{\scriptsize \ding{91}} \cite{deepseekvl2} & 0.8054  & 0.6430  & 0.8273  & 0.8361  & \mbluebf{0.6879}  & 0.8322  & 0.8399  & \mbluebf{0.7137}  & 0.8321  \\
\ding{73}Qwen3-VL (8B)\raisebox{0.5ex}{\scriptsize \ding{91}} \cite{qwen3} & 0.8377  & 0.6520  & 0.8331  & \mbluebf{0.8479}  & 0.6785  & \mbluebf{0.8481}  & \mbluebf{0.8336}  & 0.7072  & \mbluebf{0.8587}  \\
\hline
\rowcolor{gray!20}  
Ours & \mredbf{0.8687}  & \mredbf{0.7430}  & \mredbf{0.8852}  & \mredbf{0.8591}  & \mredbf{0.7243}  & \mredbf{0.8714}  & \mredbf{0.8870}  & \mredbf{0.7668}  & \mredbf{0.8959}  \\
\noalign{\vspace{-2.5pt}}
\bottomrule

\end{tabular}}
\vspace{-8mm}
\label{performances}
\end{center}
\end{table}

\begin{table*}[t!]
\belowrulesep=0pt
\aboverulesep=0pt
\centering
\renewcommand{\arraystretch}{0.78}
\small{
\caption{Comparisons of the alignment between different evaluation methods and human perception in editing models. The best results are highlighted in \mredbf{red}, and the second-best results are highlighted in \mbluebf{blue}. \raisebox{0.5ex}{\scriptsize \ding{91}} denotes fine-tuned models.} 
}
\vspace{-2mm}
\resizebox{\textwidth}{!}{
\begin{tabular}{l||c:cccc|c:cccc|c:cccc|c:c}
\toprule
\noalign{\vspace{1.5pt}}
Dimensions & \multicolumn{5}{c}{Perceptual Quality} & \multicolumn{5}{c}{Editing Alignment} & \multicolumn{5}{c}{Attribute Preservation}& \multicolumn{2}{c}{Overall Rank} \\
\cmidrule(lr){2-6}
\cmidrule(lr){7-11}
\cmidrule(lr){12-16}
\cmidrule(lr){17-18}
\noalign{\vspace{1.5pt}}
Models/Metrics & Human &\cellcolor{gray!20} Ours\raisebox{0.5ex}{\scriptsize \ding{91}} &Qwen3-VL& Q-Align\raisebox{0.5ex}{\scriptsize \ding{91}} & MANIQA\raisebox{0.5ex}{\scriptsize \ding{91}} & Human &\cellcolor{gray!20} Ours\raisebox{0.5ex}{\scriptsize \ding{91}}  &Qwen3-VL&Q-Align\raisebox{0.5ex}{\scriptsize \ding{91}}&PickScore &Human&\cellcolor{gray!20} Ours\raisebox{0.5ex}{\scriptsize \ding{91}}&Qwen3-VL&AHIQ\raisebox{0.5ex}{\scriptsize \ding{91}}&LPIPS&Human&\cellcolor{gray!20}Ours\\

\hline
\noalign{\vspace{1.5pt}}
FlowEdit(SD3) \cite{Flowedit} & 62.09 & \cellcolor{gray!20}61.25 & 90.12 & 56.79 & 0.754 & 59.65 & \cellcolor{gray!20}51.42 & 91.78 & 0.584 & 0.905 & 57.57 & \cellcolor{gray!20}51.67 & 90.12 & 0.631 & 1.924 & 1 & \cellcolor{gray!20}1 \\
RFSE \cite{RFSE}& 61.80 & \cellcolor{gray!20}61.58 & 88.52 & 58.63 & 0.790 & 55.77 & \cellcolor{gray!20}52.13 & 88.41 & 0.527 & 0.903 & 50.71 & \cellcolor{gray!20}46.67 & 85.00 & 0.569 & 3.676 & 2 & \cellcolor{gray!20}2 \\
ACE++ \cite{ACE}& 59.15 & \cellcolor{gray!20}59.29 & 88.41 & 56.08 & 0.773 & 52.87 & \cellcolor{gray!20}55.63 & 91.78 & 0.597 & 0.872 & 48.26 & \cellcolor{gray!20}41.67 & 85.04 & 0.357 & 9.697 & 3 & \cellcolor{gray!20}3 \\
CDS \cite{CDS}& 64.15 & \cellcolor{gray!20}63.83 & 95.19 & 60.88 & 0.680 & 34.50 & \cellcolor{gray!20}37.04 & 95.12 & 0.529 & 0.904 & 67.23 & \cellcolor{gray!20}60.00 & 95.15 & 0.805 & 0.215 & 4 & \cellcolor{gray!20}6 \\
FlowEdit(FLUX) \cite{Flowedit}& 56.43 & \cellcolor{gray!20}58.75 & 90.12 & 52.88 & 0.740 & 42.22 & \cellcolor{gray!20}49.38 & 90.08 & 0.583 & 0.905 & 51.18 & \cellcolor{gray!20}48.33 & 88.37 & 0.611 & 2.397 & 5 & \cellcolor{gray!20}4 \\
InfEdit \cite{InfEdit}& 51.47 & \cellcolor{gray!20}53.38 & 91.82 & 47.88 & 0.667 & 40.15 & \cellcolor{gray!20}41.96 & 93.45 & 0.512 & 0.896 & 56.22 & \cellcolor{gray!20}53.33 & 91.82 & 0.747 & 1.214 & 6 & \cellcolor{gray!20}7 \\
PnP \cite{PnP}& 56.71 & \cellcolor{gray!20}55.54 & 90.12 & 52.38 & 0.661 & 36.00 & \cellcolor{gray!20}40.63 & 95.19 & 0.539 & 0.904 & 57.41 & \cellcolor{gray!20}50.00 & 85.00 & 0.670 & 1.941 & 7 & \cellcolor{gray!20}11 \\
Any2Pix \cite{instructany2pix} & 54.31 & \cellcolor{gray!20}62.00 & 88.63 & 57.46 & 0.771 & 44.93 & \cellcolor{gray!20}53.00 & 90.23 & 0.689 & 0.833 & 44.36 & \cellcolor{gray!20}41.67 & 86.71 & 0.432 & 8.131 & 8 & \cellcolor{gray!20}5 \\
Magicbrush \cite{Magicbrush} & 52.50 & \cellcolor{gray!20}56.71 & 93.53 & 60.50 & 0.67 & 37.69 & \cellcolor{gray!20}39.00 & 93.49 & 0.503 & 0.898 & 57.13 & \cellcolor{gray!20}55.00 & 88.45 & 0.728 & 1.781 & 9 & \cellcolor{gray!20}8 \\
EDICT \cite{EDICT} & 49.67 & \cellcolor{gray!20}53.54 & 90.08 & 47.71 & 0.622 & 39.09 & \cellcolor{gray!20}41.13 & 90.04 & 0.468 & 0.857 & 55.61 & \cellcolor{gray!20}53.33 & 90.12 & 0.756 & 0.643 & 10 & \cellcolor{gray!20}9 \\
ZONE \cite{zone}& 53.40 & \cellcolor{gray!20}58.83 & 94.04 & 50.83 & 0.689 & 34.79 & \cellcolor{gray!20}36.42 & 94.00 & 0.502 & 0.906 & 59.23 & \cellcolor{gray!20}56.67 & 94.07 & 0.808 & 0.592 & 11 & \cellcolor{gray!20}10 \\
IP2P \cite{ip2p}& 49.43 & \cellcolor{gray!20}55.96 & 91.78 & 48.77 & 0.659 & 40.57 & \cellcolor{gray!20}41.71 & 91.71 & 0.460 & 0.846 & 51.18 & \cellcolor{gray!20}45.00 & 81.74 & 0.661 & 3.122 & 12 & \cellcolor{gray!20}12 \\
ReNoise \cite{renoise}& 44.80 & \cellcolor{gray!20}47.65 & 72.54 & 41.47 & 0.604 & 41.90 & \cellcolor{gray!20}43.46 & 72.50 & 0.425 & 0.826 & 48.44 & \cellcolor{gray!20}43.33 & 70.83 & 0.567 & 3.721 & 13 & \cellcolor{gray!20}14 \\
HQEdit \cite{HQ} & 45.84 & \cellcolor{gray!20}52.13 & 86.93 & 43.29 & 0.653 & 43.87 & \cellcolor{gray!20}49.96 & 88.41 & 0.500 & 0.793 & 39.77 & \cellcolor{gray!20}36.67 & 63.37 & 0.413 & 7.486 & 14 & \cellcolor{gray!20}13 \\
DDPM \cite{DDPM}& 43.90 & \cellcolor{gray!20}47.00 & 83.37 & 35.77 & 0.544 & 37.96 & \cellcolor{gray!20}39.92 & 83.37 & 0.433 & 0.631 & 48.37 & \cellcolor{gray!20}46.67 & 75.04 & 0.654 & 2.637 & 15 & \cellcolor{gray!20}15 \\
MasaCtrl \cite{Masactrl}& 41.56 & \cellcolor{gray!20}42.58 & 83.37 & 33.13 & 0.448 & 39.85 & \cellcolor{gray!20}41.92 & 78.41 & 0.486 & 0.890 & 47.14 & \cellcolor{gray!20}46.67 & 76.71 & 0.547 & 3.152 & 16 & \cellcolor{gray!20}16 \\
Text2LIVE \cite{Text2live}& 32.68 & \cellcolor{gray!20}34.02 & 83.37 & 28.90 & 0.323 & 34.16 & \cellcolor{gray!20}37.54 & 86.71 & 0.350 & 0.896 & 44.64 & \cellcolor{gray!20}43.33 & 83.37 & 0.592 & 1.733 & 17 & \cellcolor{gray!20}17 \\

\hline
\noalign{\vspace{1.5pt}}
SRCC to human $\uparrow$&&\cellcolor{gray!20}\mredbf{0.909}&0.579&\mbluebf{0.892}&0.875&&\cellcolor{gray!20}\mredbf{0.953}&0.239&\mbluebf{0.465}&0.175&&\cellcolor{gray!20}\mredbf{0.904}&0.746&\mbluebf{0.847}&0.748&&\mredbf{0.956}\\
RMSE to human $\downarrow$&&\cellcolor{gray!20}\mredbf{3.720}&37.15&\mbluebf{4.578}&51.71&&\cellcolor{gray!20}\mredbf{4.236}&47.91&42.22&\mbluebf{41.88}&&\cellcolor{gray!20}\mredbf{4.341}&\mbluebf{32.57}&51.80&49.60&&\mredbf{1.455}\\
\bottomrule
\end{tabular}

}
\label{compare_TIE}
\vspace{-4mm}
\end{table*}

\subsection{Layer Sensitivity Analysis (LSA)}
MLLM representations are hierarchical, shallow layers capture low-level patterns, while deeper layers encode abstract semantics \cite{hiprobe, headhunt}. Effective pose assessment requires a trade-off, sufficient low-level detail for artifact detection and high-level semantics for quality evaluation. We introduce LSA to profile each layer $l$ and select the optimal depth $L_{opt}$. The selection criterion $S(l)$ aggregates three normalized metrics:

\begin{itemize}
    \item \textbf{Distributional Shift ($D_{KL}$):} We measure the Kullback-Leibler divergence between the feature distributions of real ($P_{real}$) and edited ($P_{edit}$) samples. High divergence indicates strong sensitivity to manipulation:
    \begin{equation}
        D_{KL}^{(l)} = \sum_{x} P_{real}^{(l)}(x) \log \frac{P_{real}^{(l)}(x)}{P_{edit}^{(l)}(x)}
    \end{equation}
    
    \item \textbf{Class Separability (LDR):} To quantify discriminative power, we compute the Local Discriminant Ratio, defined as the ratio of between-class variance to within-class variance across feature dimensions $D$:
    \begin{equation}
        LDR^{(l)} = \frac{1}{D} \sum_{d=1}^{D} \frac{(\mu_{real, d} - \mu_{edit, d})^2}{\sigma_{real, d}^2 + \sigma_{edit, d}^2 + \epsilon}
    \end{equation}
    
    \item \textbf{Information Richness ($E$):} We calculate the Shannon entropy of feature activations ensuring the layer retains sufficient information for complex quality regression:
    \begin{equation}
        E^{(l)} = - \sum_{i} p(h_i) \log p(h_i)
    \end{equation}
\end{itemize}

The optimal layer is selected via $L_{opt} = \operatorname{argmax}_l (\hat{D}_{KL}^{(l)} + \hat{LDR}^{(l)} + \hat{E}^{(l)})$, where $\hat{\cdot}$ denotes min-max normalization. We utilize the hidden state $H^{(opt)}$ from this layer as the input for subsequent tasks.

\subsection{Task-Specific Inference}
Leveraging the optimal representation $H^{(opt)}$, we employ two decoder to handle the distinct nature of authenticity detection and quality assessment tasks.

\subsubsection{Authenticity Detection.} A multilayer perceptron maps $H^{(opt)}$ to a probability score $\hat{y}_{auth}$. This decoder is supervised by the binary cross-entropy loss $\mathcal{L}_{det}$, learning to distinguish between pristine and edited imagery.

\subsubsection{Quality Regression.} To assess perceptual and semantic quality, a multi-head regression network projects $H^{(opt)}$ into three scalar scores including perceptual quality ($\hat{s}_q$), editing alignment ($\hat{s}_e$), and attribute preservation ($\hat{s}_p$). We optimize this decoder using the Mean Squared Error (MSE) loss:
\begin{equation}
    \mathcal{L}_{qual} = \frac{1}{N} \sum_{i=1}^{N} || \mathbf{\hat{y}}_{qual}^{(i)} - \mathbf{y}_{qual}^{(i)} ||_2^2
\end{equation}
This design enables the model to correlate specific manipulation artifacts with degradation in human perceptual ratings.

\section{Experiments}

\subsection{Experimental Setup}

We utilize the InternVL3.5 backbone \cite{internvl3}. HPE-Bench is split 4:1:1 (train/val/test). We use AdamW with cosine annealing. LoRA and decoders are trained with learning rates of $1 \times 10^{-4}$ and $5 \times 10^{-5}$, respectively. Evaluation metrics include Accuracy/F1 for detection, and Spearman ($\rho_s$), Kendall ($\rho_k$), and Pearson ($\rho_p$) correlations for quality assessment.



\subsection{Authenticity Detection Performance}


\subsubsection{Comparison with Specialized Detectors}
As presented in Table~\ref{acc1}, our framework outperforms all deepfake detection methods ($\heartsuit$) and AI-generated image detectors ($\clubsuit$). Compared with the strong baseline AIDE, our method establishes new state-of-the-art performances. This advantage is observed consistently across instruction-based and description-based editing models, indicating that the learned representation is robust to different pose-editing generation mechanisms.

\subsubsection{Comparison with MLLMs}
Compared to general-purpose MLLMs ($\diamondsuit$), our framework exhibits clear performance advantages. Standard MLLMs such as Qwen3-VL and LLaVA-NeXT are substantially lower than our result. This gap demonstrates that generic multimodal representations, although effective for semantic understanding, are insufficiently sensitive to the subtle geometric and textural artifacts introduced by pose editing. In contrast, our contrastive visual tuning and LSA-based feature selection explicitly enhance manipulation-aware discrimination.

\vspace{-1mm}
\subsection{Multi-Dimensional Quality Assessment}
\vspace{-1mm}


\subsubsection{Correlation with Human Judgments}
As shown in Table~\ref{performances}, our method achieves the highest correlation with human subjective ratings across all three evaluation dimensions. Our model surpassed the leading MLLM-based metric Q-Align and the NR-IQA method MANIQA. This result indicates that the artifact-sensitive features learned for authenticity detection are also highly effective for predicting visual fidelity degradation caused by pose editing.

\subsubsection{Alignment and Preservation Analysis}
A key strength of our framework lies in its ability to capture semantic consistency between instructions and visual edits. For editing alignment, our method substantially outperforming traditional IQA metrics that are insensitive to instruction-following behavior. For attribute preservation, our model show strong capability in distinguishing intended pose modifications from unintended background or identity distortions. These results highlight the advantage of integrating artifact-sensitive visual features with semantic-awareness in unified pose-editing evaluation.

\subsubsection{Model Ranking Consistency}
To valid the practical utility of our metric, we evaluate its ability to rank different pose-editing models at the model level. As shown in Table~\ref{compare_TIE}, our method achieves the lowest RMSE and the highest correlation with human rankings. Notably, our predicted rankings consistently identify FlowEdit as the best-performing model across multiple dimensions, fully agreeing with human consensus. This demonstrates the reliability of our metric for real-world model comparison and benchmarking.

\begin{table}[t!]
\centering
\caption{Ablation study on the different backbones, decoders and LoRA tuning strategy.}
\label{ablation}
\belowrulesep=0pt
\setlength{\tabcolsep}{4pt} 
\aboverulesep=0pt
\centering
\renewcommand{\arraystretch}{1}
 \resizebox{0.48\textwidth}{!}{\begin{tabular}{lcccc|cc|cc|cc|cc}
\toprule
 \noalign{\vspace{1pt}}
\multicolumn{5}{c}{Backbone$\&$Strategy} & \multicolumn{2}{c}{Detection}& \multicolumn{2}{c}{Quality}& \multicolumn{2}{c}{Alignment}& \multicolumn{2}{c}{Preservation} \\
\cmidrule(lr){1-5} 
\cmidrule(lr){6-7}
\cmidrule(lr){8-9}
\cmidrule(lr){10-11}
\cmidrule(lr){12-13}
Backbone&Decoders&  LoRA(vision) & LoRA(llm) & LSA &  Acc$\uparrow$ & F1$\uparrow$ &  $\rho_s$$\uparrow$&  $\rho_p$$\uparrow$&$\rho_s$$\uparrow$&  $\rho_p$$\uparrow$&$\rho_s$$\uparrow$&  $\rho_p$$\uparrow$\\ 
\midrule
 \noalign{\vspace{0.5pt}}
InternVL3.5 \cite{internvl3_5} & & \checkmark &  \checkmark && 86.0 & 88.1 &0.754&0.788&0.738&0.760&0.745&0.771 \\
InternVL3.5 \cite{internvl3_5}&\checkmark &  & \checkmark & & 86.4 & 87.5 &0.780&0.803&0.764&0.780&0.746&0.789 \\
InternVL3.5 \cite{internvl3_5} & \checkmark& \checkmark &  & & 89.8 & 90.0 &0.810&0.822&0.792&0.809&0.783&0.785 \\
InternVL3.5 \cite{internvl3_5} & \checkmark& \checkmark &\checkmark  & & 91.5 & 91.8 &0.832&0.858&0.827&0.856&0.822&0.854 \\
\rowcolor{gray!20}  
InternVL3.5 \cite{internvl3_5}&\checkmark & \checkmark &  \checkmark &\checkmark & \textbf{95.5} & \textbf{91.6} &\textbf{0.868}&\textbf{0.885}&\textbf{0.859}&\textbf{0.871}&\textbf{0.887}&\textbf{0.895} \\
InternVL3 \cite{internvl3}&\checkmark & \checkmark &  \checkmark &\checkmark & 90.4 &89.8&0.827&0.849&0.837&0.866&0.858&0.860\\
DeepSeekVL2 \cite{deepseekv2} &\checkmark & \checkmark &  \checkmark &\checkmark & 87.2 &89.4 &0.805&0.827&0.836&0.832&0.839&0.832  \\
Qwen3-VL \cite{qwen3}&\checkmark & \checkmark &  \checkmark &\checkmark & 91.8 & 90.0 &0.837&0.833&0.847&0.848&0.833&0.858\\
 \noalign{\vspace{-0.5pt}}
\bottomrule
\end{tabular}}
\vspace{-2mm}
\end{table}

\subsection{Ablation Studies}

To verify the effectiveness of each component in our framework, we conducted a comprehensive ablation study using the InternVL3.5 backbone. The results are detailed in Table~\ref{ablation}.

\subsubsection{Impact of Visual Tuning}
We compare the baseline setting using only LLM LoRA and decoders against the dual LoRA setting. The introduction of contrastive visual tuning yields substantial improvement in detection accuracy and increases the quality correlation ($\rho_s$). This result confirms that standard MLLM are inherently insensitive to high-frequency editing artifacts, and our visual tuning strategy successfully injects this necessary forensic capability.

\subsubsection{Effectiveness of Layer Sensitivity Analysis}
The proposed Layer Sensitivity Analysis (LSA) provides the final and most significant performance enhancement. By comparing without LSA and the full model with LSA, we observe that the detection accuracy increases from 91.5\% to 95.5\%, accompanied by consistent improvements across all quality dimensions. This result verifies that the automatically selected intermediate layer offers a superior balance between low-level forensic cues and high-level semantic information. 

\subsubsection{Generalization Across Backbones}
To verify the universality of our framework, we applied our strategy to different backbones. Our method consistently achieves high performance on Qwen3-VL and InternVL3 backbone, demonstrating that our visual tuning and LSA mechanisms are model-agnostic and can be generalized to various MLLM backbones.

\section{Conclusion}

In this paper, we introduce HPE-Bench, a fine-grained benchmark designed for forensic supervision and instruction-aware analysis of human pose editing. Building upon this benchmark, we present a unified evaluation framework that jointly addresses authenticity detection and multi-dimensional quality assessment for fine-grained human pose editing results. Extensive experiments demonstrate that the proposed framework achieves state-of-the-art performance.



\bibliographystyle{IEEEbib}
\bibliography{icme2026references}

\end{document}